\documentclass[sigconf]{acmart}
\AtBeginDocument{%
  \providecommand\BibTeX{{%
    \normalfont B\kern-0.5em{\scshape i\kern-0.25em b}\kern-0.8em\TeX}}}

\usepackage{fontawesome}

\copyrightyear{2023}
\acmYear{2023}
\setcopyright{acmlicensed}\acmConference[MM '23]{Proceedings of the 31st ACM International Conference on Multimedia}{October 29-November 3, 2023}{Ottawa, ON, Canada}
\acmBooktitle{Proceedings of the 31st ACM International Conference on Multimedia (MM '23), October 29-November 3, 2023, Ottawa, ON, Canada}
\acmPrice{15.00}
\acmDOI{10.1145/3581783.3612336}
\acmISBN{979-8-4007-0108-5/23/10}

\usepackage{color}
\usepackage{balance}
\usepackage{multirow}
\begin{document}

\title{UniSA: Unified Generative Framework for Sentiment Analysis}


\author{Zaijing Li}
\authornote{This work was conducted when Zaijing Li was interning at Alibaba.}
\affiliation{\country{}
\institution{Central South University}
}
\email{lzj14011@gmail.com}
\author{Ting-En Lin}
\affiliation{\country{}
\institution{Alibaba Group}
}
\email{ting-en.lte@alibaba-inc.com}
\author{Yuchuan Wu}
\affiliation{\country{}
\institution{Alibaba Group}
}
\email{shengxiu.wyc@alibaba-inc.com}
\author{Meng Liu}
\affiliation{\country{}
\institution{Shandong Jianzhu University}
}
\email{mengliu.sdu@gmail.com}

\author{Fengxiao Tang}
\authornotemark[2]
\affiliation{\country{}
\institution{Central South University}
}
\email{tangfengxiao@csu.edu.cn}

\author{Ming Zhao}
\authornotemark[2]
\affiliation{\country{}
\institution{Central South University}
}
\email{meanzhao@csu.edu.cn}

\author{Yongbin Li}
\authornote{Fengxiao Tang, Ming Zhao and Yongbin Li are corresponding authors.}
\affiliation{\country{}
\institution{Alibaba Group}
}
\email{shuide.lyb@alibaba-inc.com}

\renewcommand{\shortauthors}{Zaijing Li et al.}

\begin{abstract}
Sentiment analysis is a crucial task that aims to understand people's emotional states and predict emotional categories based on multimodal information. It consists of several subtasks, such as emotion recognition in conversation (ERC), aspect-based sentiment analysis (ABSA), and multimodal sentiment analysis (MSA). However, unifying all subtasks in sentiment analysis presents numerous challenges, including modality alignment, unified input/output forms, and dataset bias. To address these challenges, we propose a Task-Specific Prompt method to jointly model subtasks and introduce a multimodal generative framework called UniSA. Additionally, we organize the benchmark datasets of main subtasks into a new Sentiment Analysis Evaluation benchmark, SAEval. We design novel pre-training tasks and training methods to enable the model to learn generic sentiment knowledge among subtasks to improve the model's multimodal sentiment perception ability. Our experimental results show that UniSA performs comparably to the state-of-the-art on all subtasks and generalizes well to various subtasks in sentiment analysis. 

%Sentiment analysis aims to uncover humans' emotional dispositions and predict emotional categories through multimodal information, which consists of many subtasks, including emotion recognition in conversation (ERC), aspect-based sentiment analysis (ABSA), multimodal sentiment analysis (MSA), etc. Unified modeling for all sentiment analysis subtasks is significant but faces many challenges, including unified I/O forms, modality alignment,  dataset bias, etc. In this paper, we recouple the various sentiment analysis subtasks and organize the benchmark datasets of the main subtasks into a new \textbf{S}entiment \textbf{A}nalysis \textbf{E}valuation benchmark, \textbf{SAEval}. To address the challenges faced in unifying multitasks for sentiment analysis, we propose Task Specifical Prompt method to jointly model subtasks and introduce a multimodal generative framework named \textbf{UniSA} to unify all multimodal subtasks as generative tasks. We design novel pre-training tasks and training methods to allow the model to learn generic sentiment knowledge among subtasks to improve the model's multimodal sentiment perception ability. Extensive experimental results demonstrate that our model performs comparably to SOTA on all subtasks and generalizes to various subtasks in sentiment analysis. In addition, we explore the bias between datasets and further analyze the reasons for the limited performance of unified modeling for sentiment analysis subtasks. 
\end{abstract}




% %% A "teaser" image appears between the author and affiliation
% %% information and the body of the document, and typically spans the

\maketitle
\begin{figure*}[t]
  \centering
  \includegraphics[width=0.95\linewidth]{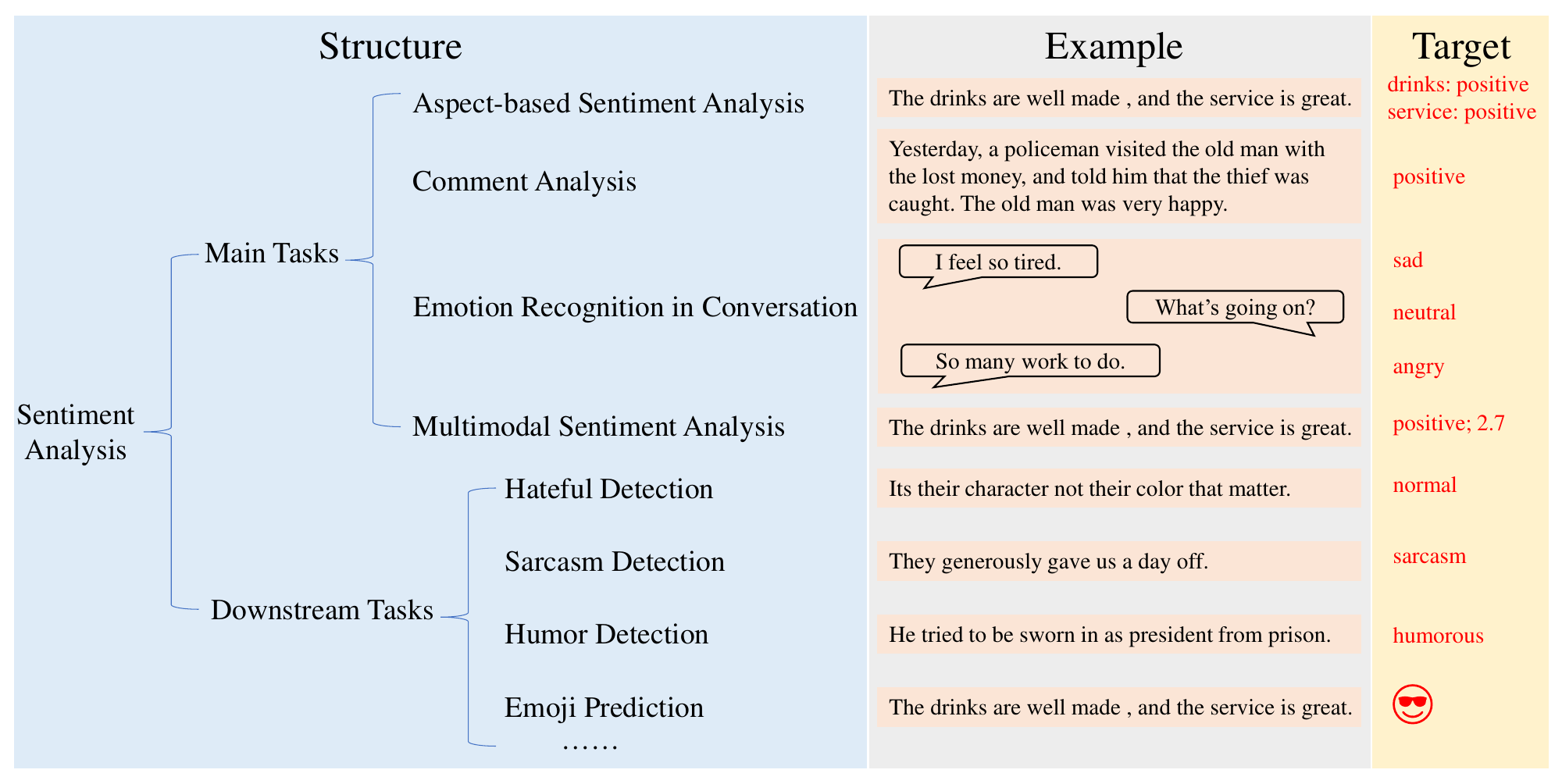}
  \caption{Subtasks of sentiment analysis are categorized into main tasks and downstream tasks based on their relevance to the emotion of humans.}
  \label{F1}
\end{figure*}
\section{Introduction}
Sentiment analysis is a discipline that leverages multimodal data to extract human opinions and comments, as well as comprehend and categorize human emotions. Generalized sentiment analysis encompasses a plethora of subtasks, such as emotion recognition in conversation (ERC), aspect-based sentiment analysis (ABSA), and multimodal sentiment analysis (MSA). Initially, research focused solely on individual subtasks; nevertheless, it has become evident that there is an interrelated sentiment knowledge among these subtasks. Hence, integrating all subtasks into a single model to enhance the sentiment understanding ability of the model has emerged as a significant objective. Following the lead of unified multi-task modeling in other domains, recent studies have explored the potential of jointly modeling some subtasks, e.g., Hu et al.~\cite{hu2022unimse} jointly modeled the ERC and MSA to boost the performance of both tasks, Yan et al.~\cite{yan2021unifiedabsa} converted all ABSA subtasks a unified generative formulation, which yielding encouraging results. Nevertheless, no research has yet been conducted on the joint modeling of all sentiment analysis subtasks (ERC, MSA, ABSA, etc.) as a single research object.

%Sentiment analysis is a study that uses multimodal information to mine humans' opinions and comments, perceive and understand humans' emotions. Generalized sentiment analysis consists of many subtasks, including emotion recognition in conversation (ERC), aspect-based sentiment analysis (ABSA), multimodal sentiment analysis (MSA), etc. Early research only focused on modeling one of these subtasks. However, there is interactive sentiment knowledge among subtasks, so modeling all subtasks to enhance the sentiment understanding ability of the model is significant. Inspired by the unified multi-task modeling in other fields, some works \cite{yan2021unifiedabsa,hu2022unimse} try to improve the model's performance on each subtask by joint multiple subtasks modeling, which has obtained inspiring results. However, they do not regard all subtasks of sentiment analysis as a whole research object. In other words, no researchers have made progress toward joint modeling of all sentiment analysis subtasks.

The unified modeling of all subtasks of sentiment analysis presents three primary challenges: 1) \textbf{Format}. The input formats and analysis views of each subtask vary. For example, MSA analyzes emotional tendencies based on single-turn utterances, ERC comprehensively assesses speaker emotions through contextual information in dialogues, and ABSA extracts attribute words from utterances and judges emotional tendencies based on those words. Jointly training these subtasks, each with different input and output formats, is the first challenge.
2) \textbf{Alignment}. Some subtasks use multimodal data (e.g., ERC), while others use single-modal data (e.g., speech emotion recognition). The data formats and representations for different modalities (text, acoustic, visual, etc.) differ, and each modality expresses emotional information in its unique way. For example, text modality uses sentiment words, attribute words, modal words, and negative words to express sentiments, while the acoustic modality mainly uses acoustic parameters such as intensity, pitch, speaking rate, and pauses to express the speaker's emotional fluctuations. The visual modality expresses human emotion through facial expressions, body posture, and eye gaze. Aligning the emotional information across modalities presents the second challenge.
3) \textbf{Bias}. Sentiment analysis is a highly subjective task, and ensuring that the model learns universal human sentiment knowledge while being less affected by subjectivity bias is the third challenge. Additionally, dataset annotation bias may affect the quality of multimodal data with high-quality annotations, making it difficult to train models that can generalize across different datasets.

In response to the challenges of sentiment analysis, we reorganize the sentiment analysis subtasks into two categories, namely main tasks and downstream tasks, based on their relevance to sentiment. As shown in Figure \ref{F1}, the main tasks, which are the subtasks most correlated with human emotional representation, include ABSA, MSA, ERC, and Comment Analysis (CA). Downstream tasks include tasks related to sentiment analysis but not necessarily detecting human emotion categories, such as irony detection, humor detection, and emoji prediction.
Meanwhile, we propose a novel multimodal sentiment analysis framework, named UniSA \footnote{The UniSA is available at : https://github.com/dawn0815/UniSA}, which takes the first step toward unified modeling of all sentiment analysis main tasks and generalizes to downstream tasks. 

Specifically, to tackle the first challenge of unifying input and output forms across different subtasks, we introduce the task-specific prompt method which treats all subtasks as generation tasks and jointly trains them. The second challenge of aligning emotional information across different modalities is addressed by extending the generative Transformer~\cite{r37} architecture to process multimodal data and proposing the modal mask training method to learn the inter-modality relationship. The third challenge of learning the difference between subtasks is tackled by introducing dataset embedding to bridge the annotation bias between different datasets. In order to evaluate the performance of our proposed framework, UniSA, we collate benchmark datasets for each subtask and construct a new sentiment analysis benchmark, SAEval, as illustrated in Table \ref{T1}. The details of this benchmark are presented in Section 3.

The contributions of this paper can be summarized as follows:
\begin{itemize}
    \item We advance a novel approach to sentiment analysis, UniSA, which unifies all subtasks under a single generative framework. This represents a significant advancement in the field, as no previous work has taken such a comprehensive approach to sentiment analysis.
    
    %\item We conduct few-shot learning experiments on various downstream sentiment analysis tasks and demonstrate excellent performance in low-resource settings. This provides evidence of the broad applicability of the UniSA model to diverse sentiment analysis tasks. Furthermore, we investigate the impact of bias between datasets and evaluate why unified modeling for sentiment analysis has limited performance.
    \item We propose novel sentiment-related pre-training tasks that allow the model to learn generic sentiment knowledge across subtasks. Extensive experimental results demonstrate that UniSA performs comparably to the state-of-the-art on all subtasks.
    \item We curate a benchmark dataset, SAEval, which comprises benchmark datasets for various sentiment analysis subtasks in a unified format, enabling comprehensive and fair evaluation of sentiment analysis models.
\end{itemize}

%1. To the best of our knowledge, our UniSA is the first work in sentiment analysis that models all subtasks in a unified way. We refine the characteristics of each subtask, design novel structures and training methods according to the characteristics of each subtask, and construct a new multimodal sentiment analysis framework, which achieves comparable performance to SOTA on all main tasks. 

%2. We conduct few shot learning experiments on various downstream tasks of sentiment analysis and obtain excellent performance in low-resource scenarios. It demonstrates the good generality of the proposed UniSA on various tasks of sentiment analysis. Going a step further, we explore the bias between datasets and analyze why unified modeling for sentiment analysis has limited performance. 

%3. We sorted out the benchmark datasets of various sentiment analysis subtasks, processed them into the same data format, and combined them into the SAEval Benchmark, which contributing to the unified research of sentiment analysis.

%The rest of the paper is organized as follows: Section 2 discusses related works; Section 3 introduces the SAEval Benchmark; Section 4 introduces the proposed UniSA framework in detail; Section 5 present the experiment setups and the analysis of experiment results; finally, Section 6 concludes the paper.
% \usepackage{multirow}

\section{Related Work}
\subsection{Sentiment Analysis}
We provide a brief overview of the recent advancements in various subfields of sentiment analysis. 
\begin{table*}[t]
\caption{Statistics of our SAEval benchmark, where T, A, and V represent text, acoustic, and visual, respectively.} %The \textcolor{red}{\faAsterisk} symbol indicates datasets used for aspect-based sentiment analysis, the \textcolor{blue}{\faChrome} symbol indicates datasets used for multimodal sentiment analysis, the \textcolor{yellow}{\faDiamond} symbol indicates datasets used for emotion recognition in conversation, and the \textcolor{green}{\faComment} symbol indicates datasets used for comment analysis.}
\label{T1}
\begin{tabular}{cccccccc}
\toprule
Task & Task type      & Dataset                     & Modality & Average Length (text) & Total Size & Train+Val Size & Test Size \\ \midrule
ABSA & Classification & SemEval-2014                & T        & 16                    & 4.8k       & 3819           & 1028      \\
ABSA & Classification & SemEval-2016                & T        & 15                    & 1.3k       & 1067           & 326       \\
MSA  & Regression     & MOSI                        & T+A+V    & 12                    & 2K         & 1513           & 686       \\
MSA  & Regression     & MOSEI                       & T+A+V    & 20                    & 20k        & 18197          & 4659      \\
ERC  & Classification & IEMOCAP                     & T+A+V    & 12                    & 7k         & 5758           & 1622      \\
ERC  & Classification & MELD                        & T+A+V    & 8                     & 13k        & 11100          & 2610      \\
ERC & Classification & DailyDialog                 & T        & 15                    & 100k       & 95239          & 7740      \\
ERC  & Classification & EmoryNLP                    & T        & 10                    & 12k        & 11278          & 1328      \\
ERC  & Classification & EmoWOZ                      & T        & 11                    & 167k       & 74983          & 8634      \\
CA   & Classification & SST-2                       & T        & 10                    & 11k        & 68221          & 1821      \\
CA   & Classification & IMDB  & T        & 233                   & 50k        & 25000          & 25000     \\
CA   & Classification & Amazon Review               & T        & 21                    & 37m        & 37m            & -         \\ \bottomrule
\end{tabular}
\end{table*}
Aspect-based sentiment analysis (ABSA) is a task that aims to identify the sentiment polarity associated with aspect terms in a single-turn utterance. Previous works have employed the attention mechanism integrated with LSTM-based neural network models to model the relationship between aspects and their contextual words~\cite{wang-etal-2016-attention,liu-zhang-2017-attention,ma2017interactive}. Recent research on ABSA has explored the application of sequence-to-sequence learning and pre-trained language models to achieve promising results~\cite{ma-etal-2019-exploring,li-etal-2020-conditional}.

Multimodal sentiment analysis involves identifying the speaker's emotion from a single-turn utterance by considering multiple modalities. Early research in this area primarily focused on geometric manipulation in feature spaces~\cite{zadeh2017tensor,zadeh2018memory}. Recent research on multimodal sentiment analysis has emphasized the importance of modal consistency and difference through multi-task joint learning~\cite{yu2021learning} or modal translation~\cite{mai2020modality}, leveraging cross-modality and multi-scale modality representation to implement the modal alignment~\cite{tsai-etal-2019-multimodal,luo2021scalevlad}.

Comment analysis involves identifying the user's emotion from one or more sentences in a comment. Recent works~\cite{xie2020unsupervised,yang2019xlnet,sachan2019revisiting}  in this field have used pre-trained language models, such as BERT~\cite{r48}, RoBERTa~\cite{r44}, and XLNet~\cite{yang2019xlnet}, to fine-tune comment datasets and achieve promising results.

Emotion recognition in conversation aims to identify the speaker's emotion from multiple utterances in a conversation \cite{lin2022duplex, qian2023empathetic, gao2023unsupervised, si2023spokenwoz}. Early research focused on context modeling using GRU~\cite{r26} models to extract context information and judge the emotion category of the utterance based on the context information~\cite{r23,r27,r28,r29,r30}. More recent research has introduced GCN models~\cite{hamilton2017inductive} into conversation scene modeling, where each utterance in the conversation \cite{zhang2022slot, lin2020discovering} is regarded as a node in the graph, and the relationship between utterances constitutes the edge connecting the nodes~\cite{r32,r33,r35,r36}. The most recent works in this area employ Transformer architecture and self-attention mechanisms to capture contextual information of utterances and achieve state-of-the-art performance in emotion recognition in conversations
~\cite{r40,r41,r43,r45,li2022emocaps}.

\subsection{Multi-task Unified Framework}
In recent years, multi-task unified architectures have shown great potential and achieved impressive results across various domains \cite{he2022galaxy, he2022space, he2022unified, yu2023speech,nie2022search,qu2021dynamic}. Bao et al.~\cite{bao2022vlmo} presented a unified vision-language pre-trained model that utilizes a modular Transformer network to jointly learn a dual encoder and a fusion encoder. Li et al.~\cite{li2021unimo} proposed a unified pre-training architecture that can effectively adapt to both single-modal and multi-modal understanding and generation tasks. Zhang et al.~\cite{zhang2022unims} proposed a Unified framework for multimodal summarization. In the Named Entity Recognition domain, some works designed unified architectures for various subtasks~\cite{yan2021unified,li2022unified}. Recently, large multimodal models such as ERNIE Bot~\cite{wang2021ernie} and GPT-4~\cite{OpenAI2023GPT4TR} have achieved remarkable results and have attracted attention from researchers in various fields.

In the sentiment analysis field, Yan et al.~\cite{yan2021unifiedabsa} employed an improved BART~\cite{r46} architecture to solve all ABSA subtasks in an end-to-end framework. Hu et al.~\cite{hu2022unimse}  proposed a multimodal sentiment knowledge-sharing framework that unifies MSA and ERC tasks from features, labels, and models. These multi-task unified works in various fields support the feasibility of unified modeling for all sentiment analysis subtasks. However, there is currently no end-to-end architecture that can model all subtasks of sentiment analysis.

%\subsection{Multimodal Pre-trained Framework}
%The architecture of multimodal pre-trained models can be classified into multiple-stream and single-stream models based on the model design. The multiple-stream architecture~\cite{lu2019vilbert,tan2019lxmert,yu2021ernie} utilizes two or more independent models to process different modalities of information, such as image, audio, and text, respectively. This architecture includes co-attention transformer layers that fuse the information from the different modalities. Examples of models that employ multiple-stream architecture are VilBERT, LXMERT, and ERNIE.

%On the other hand, single-stream architecture~\cite{li2019visualbert,VL-BERT,kim2021vilt} first fuses the information from different modalities and then inputs the fused information into the same model for processing. VisualBERT, VL-BERT, and ViLT are examples of models that utilize single-stream architecture.

%Recently, large multimodal models such as ERNIE Bot~\cite{wang2021ernie} and GPT-4~\cite{OpenAI2023GPT4TR} have achieved remarkable results and have attracted attention from researchers in various fields.

\section{SAEval: The Benchmark}
To better evaluate the performance of the model on various sentiment analysis tasks, we formalized the benchmark dataset for the main task and constructed a new benchmark, SAEval \footnote{The SAEval benchmark : https://github.com/dawn0815/SAEval-Benchmark}. This section provides a description of the datasets that constitute the SAEval benchmark and the evaluation metrics used.  
%In this section, we describe the datasets that make up the SAEval benchmark and evaluation metrics. The relevant statistical details are shown in Table \ref{T1}, \textcolor{red}{\faAsterisk} means aspect-based sentiment analysis, \textcolor{blue}{\faChrome} means multimodal sentiment analysis, \textcolor{yellow}{\faDiamond} means emotion recognition in conversation, \textcolor{green}{\faComment} means comment analysis, respectively. 

\begin{figure*}[t]
  \centering
  \includegraphics[width=1\linewidth]{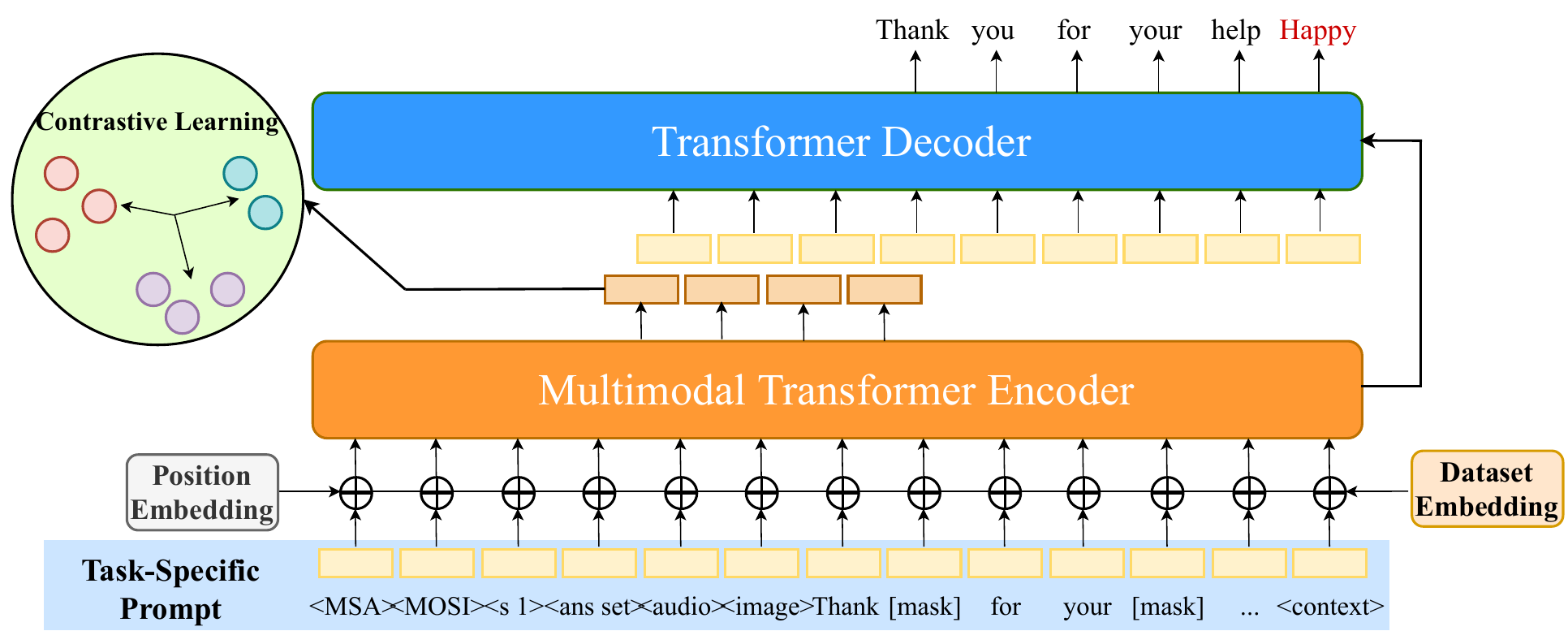}
  \caption{Overview of UniSA: a novel approach for unified sentiment analysis modeling.}
  \label{F3}
\end{figure*}
\subsection{Datasets}
As presented in Table \ref{T1}, the SAEval benchmark includes several datasets from different subtasks of sentiment analysis:
\begin{itemize}
    \item 
    SemEval-2014~\cite{pontiki-etal-2014-semeval} and SemEval-2016~\cite{pontiki2016semeval} are subtasks of the Semeval Aspect-based Sentiment Analysis challenge. The goal of these subtasks is to identify the sentiment polarity (positive, negative, neutral, conflict) corresponding to all attribute words contained in each sentence. 
    %SemEval-2014~\cite{pontiki-etal-2014-semeval} and SemEval-2016~\cite{pontiki2016semeval} are subtasks of the Semeval Aspect-based Sentiment Analysis challenge. The goal of SAEval for these two datasets is to identify the sentiment polarity (positive, negative, neutral, conflict) corresponding to all attribute words contained in each sentence.
    \item MOSI~\cite{zadeh2016mosi} and MOSEI~\cite{zadeh2018multimodal} are two widely used multimodal sentiment analysis datasets. The goal of SAEval for these two datasets is to predict the sentiment score, which is a continuous value ranging from -3 to +3, of single-turn utterances by incorporating multiple modalities.
    \item IEMOCAP~\cite{busso2008iemocap} and MELD~\cite{poria2019meld} are both datasets for emotion recognition in conversations using multimodal information. The SAEval benchmark uses these datasets to identify the emotion category of each utterance based on the multimodal information and context available.
    \item EmoryNLP~\cite{zahiri2017emotion}, DailyDialog~\cite{li2017dailydialog}, and EmoWOZ~\cite{feng2022emowoz} are datasets for textual emotion recognition in conversation. The goal of SAEval for these datasets is to identify the emotion category of utterances based on textual information and context.
    \item 
  SST-2~\cite{socher2013recursive}, IMDB~\cite{imdb}, and Amazon Review~\cite{ni2019justifying} are datasets for comment analysis. The goal of SAEval for these datasets is to identify the sentiment polarity of comments. It is important to note that Amazon Review is only used for the pre-training phase and does not have a test set for evaluation in SAEval.
\end{itemize}
%\noindent \textbf{SemEval-2014} \cite{pontiki-etal-2014-semeval} and \textbf{SemEval-2016} \cite{pontiki2016semeval}: Both datasets are subtasks of the Semeval Aspect-based Sentiment Analysis challenge. The goal of SAEval for these two datasets is to identify the sentiment polarity (positive, negative, neutral, conflict) corresponding to all attribute words contained in each sentence.

%\noindent \textbf{MOSI} \cite{zadeh2016mosi} and \textbf{MOSEI} \cite{zadeh2018multimodal}: Both datasets belong to the multimodal sentiment analysis task. The goal of SAEval for these two datasets is to identify the sentiment score (continuous numbers from -3 to +3) of single-turn utterance according to multimodal information.

%\noindent \textbf{IEMOCAP} \cite{busso2008iemocap} and \textbf{MELD} \cite{poria2019meld}: Both datasets belong to the emotion recognition in conversation task under the multimodal case. The goal of SAEval for these two datasets is to identify the emotion category of utterances according to multimodal information and context.

All datasets are unified and stored in a dictionary format \footnote{Please see Appendix A.1 for formatted samples}. The dictionary includes keywords, such as ``Task Type'', ``Dataset ID'', ``Text'', ``Audio'', and ``Image''. For ERC datasets, additional information such as ``Context'', ``Speaker ID'', and ``Utterance index'' are included to determine the conversation information of the current query.
%\noindent \textbf{EmoryNLP} \cite{zahiri2017emotion}, \textbf{DailyDialog} \cite{li2017dailydialog} and \textbf{EmoWOZ} \cite{feng2022emowoz}: These datasets belong to the textual emotion recognition in conversation task. The goal of SAEval for these datasets is to identify the emotion category of utterances according to context.

%\noindent \textbf{SST-2} \cite{socher2013recursive}, \textbf{IMDB} \cite{imdb} and \textbf{Amazon Review} \cite{ni2019justifying}: These datasets belong to the comment analysis task. The goal of SAEval for these datasets is to identify the sentiment polarity of comment. It is worth noting that Amazon Review is only used for the pre-training phase and does not have a test set for evaluation.

%We unify all datasets and store the data in dictionary format. The keywords in the dictionary contain "Task Type", "Dataset ID", "Text", "Audio" and "Image". In addition, the ERC datasets contain "Context", "Speaker ID", and "Utterance index" to determine the conversation information of the current query.

\subsection{Evaluation Metrics}
SAEval uses the same evaluation metrics as the original tasks for each subtask. Weighted Accuracy (WA) is used for aspect-based sentiment analysis and comment analysis. Mean Absolute Error (MAE), 7-category Accuracy (ACC-7), and 2-category Accuracy (ACC-2) are used for multimodal sentiment analysis. WA and weighted F1 score (WF1) are used for emotion recognition in conversation.

As the percentage of neutral categories in the DailyDialog dataset is more than 90\%, this dataset is usually measured with neutral categories removed by default, and macro-averaged F1 (MF1) is used as the measure.

%We use the same evaluation metric from the original tasks: Weighted Accuracy (WA) for aspect-based sentiment analysis and comment analysis; Mean Absolute Error (MAE), 7-category Accuracy (ACC-7) and 2-category Accuracy (ACC-2) for multimodal sentiment analysis; WA and weighted F1 score (WF1) for emotion recognition in conversation. Due to the percentage of neutral categories in the DailyDialog dataset is more than 90\%, this dataset is usually measured with neutral categories removed by default, and macro-averaged F1 (MF1) is used as the measure.

\section{Methodology}
%In this section, we propose novel training methods and framework to address the challenges mentioned above of unified multi-task modeling for sentiment analysis, and design sentiment-related pre-training tasks to improve the generic sentiment perception of the models.
The architecture of our UniSA is depicted in Figure \ref{F3}. We adopt a generative Transformer architecture to unify all subtasks of sentiment analysis into generation tasks. Concretely, to handle the cross-modality inputs of visual, acoustic, and text, we modify the original Transformer encoder to a multimodal encoder and introduce a Modal Mask Training method. This method enables the model to learn the relationship between different modalities effectively. We also propose a Task-Specific Prompt method to standardize the input format of all subtasks. Furthermore, to address the bias between datasets, we incorporate a dataset embedding in the input to differentiate between different datasets. This technique helps the model to better understand the characteristics of each dataset and improves its performance on all tasks. We will elaborate on each of them sequentially.
\subsection{Problem Formulation}
The unified sentiment analysis subtasks aim to process arbitrary data from different modalities, such as text, acoustic, and visual, and output emotion predict results corresponding to a specific subtask. The subtask set includes ABSA, MSA, ERC, and CA, while the modality set includes $m_t$, $m_a$, and $m_v$, corresponding to text, acoustic, and visual modalities, respectively. The task involves both multimodal situations, such as MSA and ERC, as well as unimodal situations, such as ABSA and CA. Moreover, the task must also consider multimodal situations where one or more modalities are missing.
\begin{table*}[t]
\caption{Number of model parameters (in millions) and hyperparameter settings for the fine-tuning phase.}
\label{T2}
\begin{tabular}{ccccccc}
\toprule
Backbone    & Number of Parameters & Learning Rate & Batch Size & Dropout Rate & Epochs & Max Length of Sequence \\ \midrule
GPT2-medium & 341m  & 5e-5          & 32         & 0.0          & 15     & 128                    \\
T5-base   & 221m  & 5e-5          & 32         & 0.0          & 15     & 128                    \\
BART-base   & 141m & 5e-6          & 64         & 0.1          & 40     & 600                    
\\ \bottomrule
\end{tabular}
\end{table*}
Our target is to build a multimodal emotion-aware framework based on SAEval, named UniSA, which learns cross-task emotional knowledge through emotion-related pre-training tasks and generalizes to various downstream tasks. 

%Given a subtasks set $\{ABSA, MSA, ERC, CA\}$ and a modality set $\{m_t, m_a, m_v\}$,  where $t$, $a$, $v$ denote the text, acoustic and visual. The goal of unified sentiment analysis subtasks is to input arbitrary subtask data under different modal settings and output the results corresponding to a specific subtask. The task involves multimodal situations (including MSA, ERC, etc.) and unimodal situations (including ABSA, CA, etc.). In addition, the task must consider multimodal situations where some modalities are missing.
%\subsection{Our Framework}

%Figure \ref{F3} illustrates the architecture of our UniSA. We introduce a generative Transformer architecture to unify all sentiment analysis subtasks into generation tasks, and propose a Task Specifical Prompt method to normalize the input form of all subtasks. In order to adapt the model to cross-modality inputs of visual, acoustic, and text, we modify the original Transformer encoder to a multimodal encoder and propose a Modal Mask Training method so that the model can thoroughly learn the relationship between different modalities. In addition, considering the bias between datasets, we introduce dataset embedding in the input to distinguish the different datasets.

\begin{figure*}[h]
  \centering
  \includegraphics[width=\linewidth]{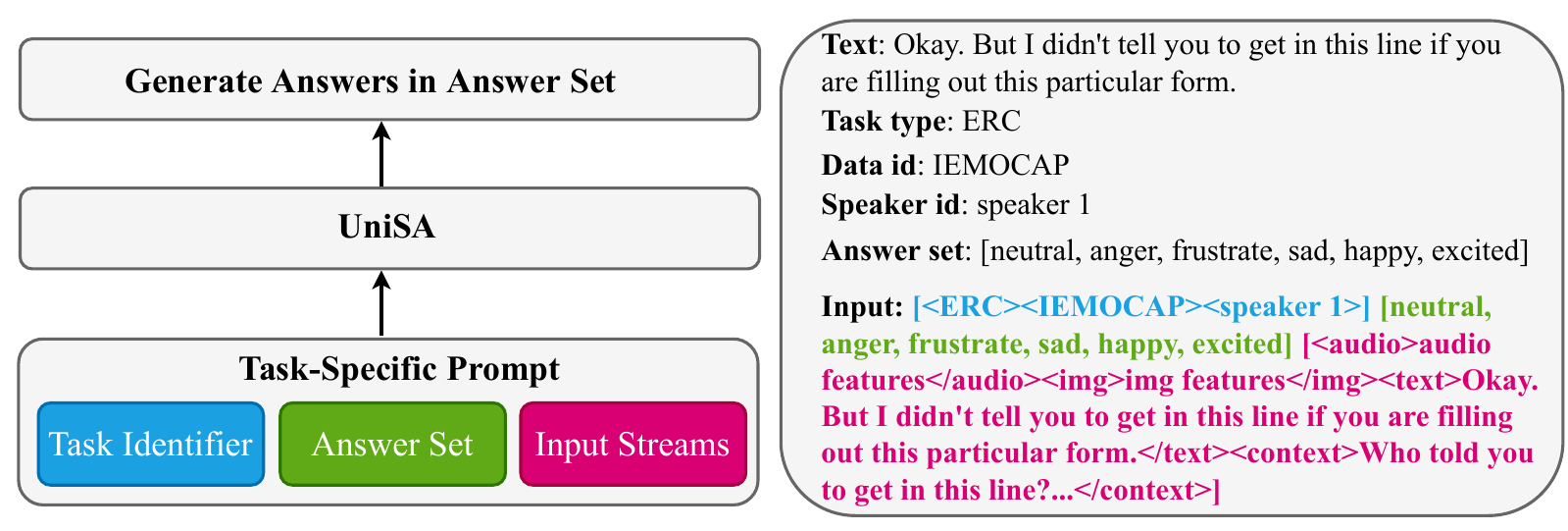}
  \caption{An example of our Task-Specifical Prompt method.}
  \label{F4}
\end{figure*}

%In each subtask of sentiment analysis, text modality data is abundant, while multimodality data is limited. To improve the multimodal emotion perception capability of the model, we propose a Modal Mask Training method. Specifically, given a multimodal input signal $I_i = \{I_{i}^{t}, I_{i}^{a}, I_{i}^{v} \}$, where $I_{i}^{m}, m\in \left \{ t, a, v \right \} $ represent unimodal input of video fragment $i$, we force the masking of one or more of the modal inputs so that multimodal data is transformed into seven modal settings: $\{  I_{i}^{t}, I_{i}^{a}, I_{i}^{v}, I_{i}^{a} + I_{i}^{v}, I_{i}^{t} + I_{i}^{a}, I_{i}^{t} + I_{i}^{v}, I_{i}^{t} + I_{i}^{a} + I_{i}^{v} \} $. It is worth noting that some modal settings do not work well, so we only use the following settings during the training phase: $\{ I_{i}^{t}, I_{i}^{t} + I_{i}^{a}, I_{i}^{t} + I_{i}^{v}, I_{i}^{t} + I_{i}^{a} + I_{i}^{v} \} $.

\subsection{Task-Specific Prompt}
To jointly model different subtasks, we propose Task-Specific Prompt to unify the input streams of all subtasks, and transform all subtasks into generative tasks to unify the output form of subtasks. Task-Specific Prompt comprises three components: task identifier $Z$, answer set $Y$, and input streams $X$. The template $L$ can be represented as:
\begin{equation}
  L =  \{ Z, Y, X  \}. 
\end{equation}

As illustrated in Figure \ref{F4}, the task identifier $Z$ is made up of special tokens, including $<task\_type>$, $<data\_id>$, $<speaker\_id>$, etc., which distinguish different subtasks, datasets, and speakers (in conversation). The answer set $Y$ is a specific set of labels for each dataset that guides the model in generating the expected results. The input streams $X$ indicate the text, acoustic, visual, and context of the inputs. Task-Specific Prompt standardizes the input form and guides the model to generate task-specific results according to the answer set.

%Task Specifical Prompt consists of three parts: task representation $P$, answer sets $A$, and input streams $X$. So the template T could be expressed as:

%As shown in Figure \ref{F4}, the task representation $T$ consists of special tokens, including $<task\_type>$, $<data\_id>$, $<speaker\_id>$, etc., to distinguish different subtasks, datasets, and speakers (in conversation). The answer set $A$ is a specific set of labels for each dataset, to guide the model in generating the expected results. The input streams $X$ indicate the text, acoustic, visual and context of the inputs. 
\subsection{Modal Mask Training}
As the text modality data is abundant in each subtask of sentiment analysis, while the multimodal data is limited, we propose a Modal Mask Training method to enhance the model's multimodal emotion perception capability. Specifically, given a multimodal input signal $I_i = {I_{i}^{t}, I_{i}^{a}, I_{i}^{v} }$, where $I_{i}^{m}, m\in \{ t, a, v \} $ represent unimodal input of time $i$, we force the masking of one or more of the modal inputs. This transformation yields seven modal settings: ${ I_{i}^{t}, I_{i}^{a}, I_{i}^{v}, I_{i}^{a} + I_{i}^{v}, I_{i}^{t} + I_{i}^{a}, I_{i}^{t} + I_{i}^{v}, I_{i}^{t} + I_{i}^{a} + I_{i}^{v} } $.

However, we found that some modal settings do not work well during training, so we only use the following four modal settings : ${ I_{i}^{t}, I_{i}^{t} + I_{i}^{a}, I_{i}^{t} + I_{i}^{v}, I_{i}^{t} + I_{i}^{a} + I_{i}^{v} } $. In this way, multimodal data has four different input forms in the training phase, which extends the modal diversity of the training data and can effectively cope with the absence of some modalities in multiple data in real scenarios. This method allows the model to learn the relationship between different modalities effectively and improve its performance on multimodal sentiment analysis tasks where the data is limited.
\subsection{Dataset Embedding}
To reduce the impact of dataset bias on the model performance, we propose the use of dataset embedding. This technique transforms the dataset indexes into one-hot embeddings, allowing the model to distinguish between different datasets. As sentiment analysis is a subjective task, people's emotional responses to the same sentence can vary. Additionally, different datasets may be labeled by different annotators, introducing subjective bias between datasets. The dataset embedding helps mitigate this bias, allowing the model to better generalize across different datasets.

%Sentiment analysis is a subjective task, and people's emotional response to the same sentence varies, i.e., "There are a thousand Hamlets in a thousand people's eyes." Further, different datasets are labeled by different annotators, so there is also subjective bias between datasets. To reduce the impact of dataset bias on model performance, we propose a new embedding, namely dataset embedding, which transforms dataset indexes into one-hot embeddings to distinguish different datasets.
\subsection{Multimodal Transformer Encoder}
The encoder of our model is based on a multi-layer bidirectional Transformer, which is similar to the architecture used by Xing et al. \cite{xing2021km}. We adopt a single-stream architecture to jointly train the inputs of different modalities. This architecture allows the model to effectively capture the interactions and dependencies between different modalities and improve its overall performance on all tasks. For visual modality, we use pre-trained MobileNet \cite{howard2017mobilenets} extracted features as visual embedding, for acoustic modality, we use librosa \cite{mcfee2015librosa} extracted fbank features as acoustic embedding; and for text modality, we follow the setting of the BART model~\cite{r46}.
\subsection{Pre-training Tasks}
To improve the emotional perception ability of the model, we propose novel pre-training tasks and divide the pre-training into two stages to guide the model to learn general emotional knowledge gradually.

\subsubsection{Mask Context Modeling}
We extend the Masked Language Modeling \cite{r48} to a multimodal input and context mask for the current query, named Mask Context Modeling (MCM).MCM can randomly mask tokens from the input text, acoustic, visual, and context modalities to encourage the model to learn to predict the masked tokens. To promote better generalization, we increase the mask probability to 50\%. We denote the mask indices by $1 \leq m \leq M$, where $M$ is the number of masked tokens. We denote the masked token by $w_m$, and the remaining tokens that are not masked by $w_n$. The loss function for MCM is defined as:
\begin{equation}
\mathcal{L}_{MCM}(\theta ) = -\sum_{m=1}^{M} \log_{}{(P_{\theta }(w_{m} | w_{n} ) )}, 
\end{equation}
where $P_{\theta }$ denotes the output distribution of the model, and $\theta$ represents model parameters to be optimized.

\subsubsection{Sentiment Polarity Prediction}
To encourage the model to learn to distinguish between different sentiment categories, we transform the fine-grained emotion labels of the datasets into sentiment polarity labels by mapping them to positive, negative, and neutral categories. Then, in the Sentiment Polarity Prediction (SPP) task, we train the model to predict the sentiment polarity category of the input. The loss function for SPP is defined as:
%We transformed the fine-grained emotion labels of the datasets into sentiment polarity labels by label mapping. Then, we denote the sentiment polarity set $s \in \{ positive, negative, neutral \}$. In the Sentiment Polarity Prediction (SPP), we predict the sentiment polarity category of the input, and the the loss function for SPP is defined as:
\begin{equation}
\mathcal{L}_{SPP}(\theta ) = - \log_{}{(P_{\theta }(s | w_{n} ) )}. 
\end{equation}
\subsubsection{Coarse-grained Label Contrast Learning}
To encourage the model to learn to distinguish between different sentiment categories at a coarse level, we propose the Coarse-grained Label Contrast Learning (CCL) task. In this task, we take the output of the encoder as a representation of each sample and compute the Euclidean distance between samples with the same sentiment label in a batch. We then maximize the similarity between samples with the same sentiment label. The loss function for CCL is defined as:
%The goal of Coarse-grained Contrast Learning (CCL) is to maximize the similarity between samples with the same sentiment label. In the CCL task, we take the output of the encoder as a representation of each sample and compute the Euclidean distance between samples with the same sentiment label in a batch samples, and the loss function for CCL is defined as:
\begin{equation}
\mathcal{L}_{CCL}(\theta ) = \sum_{j=1}^{b} \frac{\sum_{k=1}^{b}(distance(j,k)*mask(j,k)) }{\sum_{k=1}^{b}  distance(j,k)} ,
\end{equation}
where $b$ denotes the batch size, $distance(j,k)$ denotes the Euclidean distance between sample $j$ and sample $k$, and $mask(j,k)$ is 1 when sample $j$ and sample $k$ have the same sentiment label, and 0 otherwise.

\subsubsection{Cross-task Emotion Prediction}
In the Cross-task Emotion Prediction (CEP) task, we first take the output of the encoder as the representation of each sample and cluster the samples for each subtask. Then, for each sample, we calculate the distance between its representation and each label cluster of each subtask. We take the label corresponding to the cluster with the smallest distance as the pseudo label of the sample for each subtask. Given a subtask set $D =\{ABSA, MSA, ERC, CA\}$, each sample will have four labels: one for the original label and three for the cross-task pseudo labels. The loss function for CEP is defined as:
%In the Cross-task Emotion Prediction (CEP) task, we first take the output of the encoder as the representation of each sample, and cluster the samples for each subtask, then calculate the distance between sample $j$ and each label cluster of subtask $d$, and take the label corresponding to the cluster with the smallest distance as the pseudo label of sample $j$ on subtask $d$. Given a subtask set $D \in \{ABSA, MSA, ERC, CA\}$, each sample will have four labels: one for the original label, and three for the cross-task pseudo labels. The loss function for CEP is defined as:
\begin{equation}
\mathcal{L}_{CEP}(\theta ) = - \sum^{D} \log_{}{(P_{\theta }(E_d | w_m ) )}, 
\end{equation}
where $E_d$ denotes the emotion label in subtask $d \in  D$.

\subsubsection{Pre-training Stage One}
The first stage is coarse-grained emotion perception pre-training, including Mask Context Modeling, Sentiment Polarity Prediction, and Coarse-grained Label Contrast Learning. It aims to allow the model to acquire preliminary sentiment classification capabilities. Inspired by the Directional Expectation Test \cite{ribeiro2020beyond}, we argue that emotions are polarity-invariant: combining queries with the same sentiment polarity does not change both. Therefore, we divide all datasets into different data pools according to sentiment polarity. Then, we randomly select two queries from the same data pool to combine into a new query for the pre-training stage 1. The loss function for pre-training stage one is defined as:
%The first stage is coarse-grained emotion perception pretraining, including Mask Context Modeling, Sentiment Polarity Prediction, and Coarse-grained Label Contrast Learning. It aims to allow the model to acquire preliminary sentiment classification capabilities. Inspired by the Directional Expectation Test \cite{ribeiro2020beyond}, we argue that emotions are polarity invariant: combining queries with the same sentiment polarity does not change both. Therefore, we divide all datasets into different data pools according to sentiment polarity. Then we randomly select two queries from the same data pool to combine into a new query for the pre-training stage 1. The loss function for pretraining stage 1 is defined as:
\begin{equation}
\mathcal{L}_{stage1} = \mathcal{L}_{MCM} + \mathcal{L}_{SPP} + \mathcal{L}_{CCL}.
\end{equation}

\begin{table*}[t]
\caption{Experimental results on the SAEval benchmark. $\star$ indicates the model with the pre-training phase.}
\label{T3}
\begin{tabular}{ccccccccc}
\hline
Task                  & Dataset      & Metric & SOTA Models              & SOTA Scores & $UniSA_{GPT2}$ & $UniSA_{T5}$ & $UniSA_{BART}$ & $UniSA_{BART}^{\star }$ \\ \hline
\multirow{2}{*}{ABSA} &  SemEval14                & WA $\uparrow$ & InstructABSA             & 88.37       & -           & -         & 73.61       & 82.61       \\
                      & SemEval16                & WA $\uparrow$   & InstructABSA             & 94.02       & -           & -         & 76.15       & 80.35       \\ \hline
\multirow{6}{*}{MSA}  & \multirow{3}{*}{MOSI}    & MAE $\downarrow$    & \multirow{3}{*}{UniMSE}  & 0.691       & 1.41        & 0.9004    & 0.9989      & 0.7422      \\
                      &                          & ACC-7 $\uparrow$  &                          & 48.68       & 15.45       & 37.46     & 38.92       & 48.54       \\
                      &                          & ACC-2  $\uparrow$ &                          & 85.85       & 44.75       & 76.82     & 77.11       & 84.11       \\
                      & \multirow{3}{*}{MOSEI}   & MAE $\downarrow$     & \multirow{3}{*}{UniMSE}  & 0.523       & 0.8384      & 0.5460    & 0.5720      & 0.5866      \\
                      &                          & ACC-7 $\uparrow$ &                          & 54.39       & 41.36       & 52.50     & 50.91       & 50.03       \\
                      &                          & ACC-2 $\uparrow$  &                          & 85.86       & 71.02       & 84.22     & 85.57       & 84.93       \\ \hline
\multirow{7}{*}{ERC}  & \multirow{2}{*}{MELD}    & WA $\uparrow$    & \multirow{2}{*}{SPCL-CL-ERC}  & -       & 48.12       & 64.52     & 62.45       & 62.34       \\
                      &                          & WF1 $\uparrow$   &                          & 67.25       & 31.26       & 62.17     & 60.78       & 62.22       \\
                      & \multirow{2}{*}{IEMOCAP} & WA $\uparrow$    & \multirow{2}{*}{EmoCaps} & 70.56       & 23.67       & 62.51     & 65.04       & 64.24       \\
                      &                          & WF1 $\uparrow$    &                          & 71.77       & 9.06        & 62.70     & 65.21       & 64.46       \\
                      & EmoryNLP                 & WF1 $\uparrow$   & SPCL-CL-ERC                 & 40.94       & 10.93       & 33.48     & 32.93       & 34.95       \\
                      & DailyDialog              & MF1 $\uparrow$    & CoMPM                    & 60.34       & -           & 59.38     & 58.36       & 59.24       \\
                      & EmoWoz                   & WF1 $\uparrow$   & ContextBert              & 79.7        & 57.19       & 87.70     & 90.33       & 90.52       \\ \hline
\multirow{2}{*}{CA}   & SST-2                    & WA $\uparrow$   & T5-11B                   & 97.5        & -           & -         & 91.85       & 90.71       \\
                      & IMDB                     & WA $\uparrow$   & XLNet                    & 96.21       & -           & -         & 93.35       & 92.26       \\ \hline
\end{tabular}
\end{table*}

\subsubsection{Pre-training Stage Two}
The second stage is fine-grained emotion perception pre-training, including Mask Context Modeling and Cross-task Emotion Prediction. It aims to allow the model to acquire fine-grained emotion classification capability. The loss function for pre-training stage 2 is defined as:
%The second stage is fine-grained emotion perception pre-training, including Mask Context Modeling and Cross-task Emotion Prediction. It aims to allow the model to acquire fine-grained emotion classification capability. The loss function for pretraining stage 2 is defined as:
\begin{equation}
\mathcal{L}_{stage2} = \mathcal{L}_{MCM} + \mathcal{L}_{CEP}.
\end{equation}
\section{Experiments}

\subsection{Baseline}
In this section, we reported the current SOTA models for each dataset in the SAEval benchmark, which we use as baselines to compare with the performance of UniSA on each subtask. These SOTA models are described as follows:
\begin{table*}[]
\caption{Results of ablation experiments on MOSI, MOSEI, IEMOCAP, and MELD.}
\label{T4}
\begin{tabular}{c|ccc|ccc|cc|cc}
\hline
       & \multicolumn{3}{c|}{MOSI}    & \multicolumn{3}{c|}{MOSEI}   & \multicolumn{2}{c|}{MELD} & \multicolumn{2}{c}{IEMOCAP} \\
        & MAE $\downarrow$     & ACC-7 $\uparrow$  & ACC-2 $\uparrow$      & MAE $\downarrow$    & ACC-7 $\uparrow$ & ACC-2 $\uparrow$       & WA $\uparrow$        & WF1 $\uparrow$          & WA $\uparrow$           & WF1 $\uparrow$          \\ \hline
SOTA Scores   & 0.691  & 48.68 & 85.85 & 0.523  & 54.39 & 85.86 & 67.85       & 66.71       & 70.56        & 71.77        \\
S             & 1.79   & 20.69 & 44.75 & 1.006  & 41.31 & 71.02 & 46.59       & 31.37       & 20.22        & 60.27        \\
S+P           & 1.06   & 28.86 & 69.67 & 0.6226 & 47.82 & 84.71 & 61.72       & 59.49       & 60.11        & 59.64        \\
S+P+Text only & 0.9192 & 41.39 & 78.86 & 0.6232 & 47.77 & 82.20 & 63.90       & 62.32       & 56.47        & 53.06        \\
S+P+F         & 0.9439 & 29.59 & 74.63 & 0.5712 & 52.58 & 84.88 & 66.20       & 64.19       & 63.74        & 63.60        \\
S+P+F+C       & 0.9034 & 41.25 & 78.13 & 0.5706 & 52.65 & 85.85 & 63.83       & 63.12       & 57.45        & 56.68        \\
E+P+F+C       & 0.8908 & 40.52 & 77.55 & 0.5716 & 52.17 & 85.16 & 63.98       & 63.45       & 54.87        & 53.68        \\
S+P+F+T1      & 0.8527 & 42.85 & 78.42 & 0.5544 & 53.12 & 85.81 & 62.60       & 61.03       & 57.15        & 56.97        \\
S+P+F+T2      & 0.9441 & 37.17 & 78.71 & 0.5436 & 52.62 & 85.59 & 62.64       & 61.14       & 65.10        & 64.67        \\
S+P+F+T1+T2   & 0.7422 & 48.54 & 84.11 & 0.5866 & 50.03 & 84.93 & 62.34       & 62.22       & 64.24        & 64.46        \\ \hline
\end{tabular}
\end{table*}
\begin{itemize}
    \item \textbf{UniMSE} \cite{hu2022unimse}: UniMSE is a multimodal sentiment knowledge-sharing framework that unifies MSA and ERC tasks from features, labels, and models. It is the current SOTA model for \textbf{MOSI} \cite{zadeh2016mosi} and \textbf{MOSEI} \cite{zadeh2018multimodal}.
    \item \textbf{EmoCaps} \cite{li2022emocaps}: EmoCaps is a multimodal framework for conversational emotion recognition, which proposes the multimodal emotion vector to characterize the emotional tendencies of the utterance itself. It is the current SOTA model for \textbf{IEMOCAP} \cite{busso2008iemocap}.
    \item \textbf{SPCL-CL-ERC} \cite{song-etal-2022-supervised}: SPCL-CL-ERC is a model that combines Supervised Prototypical Contrastive Learning and curriculum learning strategy to address imbalanced classification problem in conversational emotion recognition. It is the current SOTA model for \textbf{EmoryNLP} \cite{zahiri2017emotion} and \textbf{MELD} \cite{poria2019meld}.
    \item \textbf{CoMPM} \cite{lee2022compm}: CoMPM is a model for conversational emotion recognition, which combines the speaker’s pre-trained memory with the context model and finds that the pre-trained memory significantly improves the performance of the context model. It is the current SOTA model for \textbf{DailyDialog} \cite{li2017dailydialog}.
\end{itemize}

In addition, the pre-trained models fine-tuned on specific datasets have achieved promising results: \textbf{BERT} \cite{r48} for \textbf{EmoWoz} \cite{feng2022emowoz}; \textbf{T5} \cite{raffel2020exploring} for \textbf{SST-2} \cite{socher2013recursive}; \textbf{XLNet} \cite{yang2019xlnet} for \textbf{IMDB} \cite{imdb}; \textbf{InstructABSA} \cite{yang2019xlnet} for \textbf{SemEval-2014} \cite{pontiki-etal-2014-semeval} and \textbf{SemEval-2016} \cite{pontiki2016semeval}.

\begin{table*}[t]
\caption{Experimental results on the downstream tasks in terms of MF1. The SOTA models learn on the entire training set, while T5-base and UniSA are under the few-shot setting.}
\label{T5}
\begin{tabular}{cll|ccclclcl}
\hline
\multicolumn{3}{c|}{}                  & Hateval  & Twitter emoji & \multicolumn{2}{c}{Twitter emotion} & \multicolumn{2}{c}{Twitter sentiment} & \multicolumn{2}{c}{Sarcasmania} \\ \hline
\multicolumn{3}{c|}{Samples (few-shot/train)} & 300/9,000 & 3,000/4,5000    & \multicolumn{2}{c}{597/3,257}        & \multicolumn{2}{c}{450/45,615}         & \multicolumn{2}{c}{300/27,846}   \\
\multicolumn{3}{c|}{SOTA (train)}             & 65.10    & 32.20         & \multicolumn{2}{c}{76.1}            & \multicolumn{2}{c}{72.07}             & \multicolumn{2}{c}{-}           \\
\multicolumn{3}{c|}{T5 (few-shot)}            & 47.94    & 7.58          & \multicolumn{2}{c}{59.56}           & \multicolumn{2}{c}{66.13}             & \multicolumn{2}{c}{99.31}       \\
\multicolumn{3}{c|}{UniSA (few-shot)}         & 49.61    & 14.82         & \multicolumn{2}{c}{65.98}           & \multicolumn{2}{c}{64.17}             & \multicolumn{2}{c}{99.78}       \\ \hline
\end{tabular}
\end{table*}

%\subsection{Finetune on Main Tasks}
%To verify the performance of the model on the main tasks of sentiment analysis, we take the pre-trained UniSA, which is passed through two pre-training phases, and jointly fine-tune the datasets on the SAEval. 

%When making multi-task joint fine-tuning, the model tends to overfit on some tasks and underfit on others due to 1) the different learning difficulty of the subtasks, i.e., the model does not learn at the same rate across tasks, and 2) the direction of the gradient between subtasks is not consistent.

%To solve this problem, we introduce \textbf{task-average sampling} in the fine-tuning phase, so that the model learns the gradient of each task relatively during each iteration. Specifically, we divide the dataset into different task pools according to the task type, shuffle for all task pools, and then equally distribute the number of samples from each task pool for each step.
\subsection{Implementation Details}
We explored three generative architectures, GPT-2 \cite{radford2019language}, T5 \cite{raffel2020exploring}, and BART \cite{r46}, as the backbone of our UniSA, and determined the optimal model through comparative experiments. The acoustic and visual representations have a hidden dimension of 64, while the textual embedding size is 768. The batch size is set to 64, and the learning rate is 5e-6 for BART-base, while it is 5e-5 for T5-base and GPT2-medium. Further details can be found in Table \ref{T2}.

To evaluate the performance of the model on the primary tasks of sentiment analysis, we took the pre-trained UniSA, which undergoes two pre-training phases and conducted joint fine-tuning on the datasets in SAEval. During multi-task joint fine-tuning, the model may overfit on some tasks and underfit on others due to the varying learning difficulties across subtasks and inconsistent gradient directions between subtasks. To mitigate this issue, we proposed \textbf{task-average sampling} during the fine-tuning phase, which enables the model to learn the gradient of each task equally during each iteration. Specifically, we split the dataset into various task pools based on the task type, shuffled them, and distributed the same number of samples from each task pool equally for each step.

\subsection{Performance Analysis}
The experimental results of our UniSA and the existing SOTA models on the SAEval benchmark are presented in Table \ref{T3}, where vacant cells indicate that the models were not fine-tuned on the corresponding datasets. We initially fine-tuned all datasets without pre-training stages using GPT2, T5, and BART as backbones. The results show that GPT-2 with a multimodal encoder ($UniSA_{GPT2}$) has limited performance. Furthermore, we observed that BART ($UniSA_{BART}$) outperforms T5 ($UniSA_{T5}$) in extending to multimodal architectures. 
\textbf{Therefore, we chose pre-trained BART as the backbone for our UniSA model in further experiments.}

As shown in Table \ref{T3}, our proposed UniSA model performs comparably to the existing SOTA models for each dataset. Although UniSA is unable to outperform existing SOTA models on various benchmark datasets, it demonstrates the feasibility of uniformly modeling all sentiment analysis subtasks. Moreover, constrained by specific tasks and modalities, these SOTA models cannot be effectively generalized to other subtasks, whereas UniSA is an all-in-one model that can perform all sentiment analysis subtasks with a relatively small number of parameters.

%The experimental results of our UniSA with the existing SOTA model on the SAEval benchmark are shown in Table \ref{T3}. We first employ pre-trained GPT2, T5, and BART as the backbone to fine-tune all datasets without pretraining stages. The experimental results show that the performance of GPT-2 with a multimodal encoder is limited. In addition, we find that the BART model outperforms the T5 model in extending to multimodal architectures, so we choose the pre-trained BART as the backbone of UniSA for further experiments.

%As shown in Table \ref{T3}, our proposed UniSA model performs closely to the existing SOTA models for each dataset. Although UniSA fails to outperform SOTA in all benchmark datasets, it demonstrates the feasibility of uniformly modeling all sentiment analysis subtasks. Moreover, it is an all-in-one model that can perform all sentiment analysis subtasks with a small number of parameters.

\subsection{Ablation Study}
We used BART as the backbone of UniSA and conducted a series of ablation studies on the MOSI, MOSEI, IEMOCAP, and MELD datasets. Our ablation experiments aimed to investigate the effectiveness of our proposed methods and are as follows: 1) We replaced the proposed task specifical prompt method with task tokens to verify its effects on model performance. 2) We removed the modal mask training method to verify its impact on model performance. 3) To explore the impact of different input forms on model performance, we introduced three additional LSTMs as encoders for audio, image, and context inputs. 4) We validated the effectiveness of each pre-training stage. 5) We eliminated the acoustic and visual modalities from the multimodal signals to investigate their effects on model performance. 

The experimental results are shown in Table \ref{T4}, where S denotes the sequences of all modalities input to a single encoder, while E denotes extended LSTMs as encoders for acoustic and visual modalities. P denotes Task-Specific Prompt, F denotes Modal Mask Training, C denotes additional LSTMs as an encoder for context, and T1 and T2 denote the first and second stages of pre-training, respectively. The results of the ablation experiments demonstrate the effectiveness of our proposed methods.

%We conducted a series of ablation studies on the MOSI, MOSEI, IEMOCAP, and MELD. Our ablation experiments are as follows: 1) We use task tokens instead of the proposed prompt method to verify the Task Specifical Prompt effects on model performance. 2) We eliminate Modal Mask Training method to verify the effect of the method on model performance. 3) To explore the impact of audio, image, and contextual input forms on model performance, we introduce three additional LSTMs to process these three input sequences. 4) We validate the effectiveness of each pre-training stage. 5) We eliminate acoustic and visual modality from multimodal signals to verify the modal effects on model performance. The experimental results are shown in Table \ref{T4}, where S denotes sequences of all modalities input to a single encoder, while D denotes additional LSTMs as encoders for acoustic and visual modalities. P denotes Task Specifical Prompt, M denotes Modal Mask Training, L denotes additional LSTM as an encoder for context, and T1 and T2 denote the first and second stages of pre-training, respectively. The results of the ablation experiments demonstrate the effectiveness of our proposed methods.

\subsection{Few-shot Generalization}
To demonstrate the generalizability of our proposed UniSA on various subtasks, we conducted experiments on several downstream tasks under the few-shot setting. We reported the SOTA scores of several downstream datasets, such as the SemEval2019 Hateval challenge \cite{basile-etal-2019-semeval}, Sarcasmania \cite{kocon2023chatgpt}, Semeval2017 Sentiment Analysis Challenge \cite{rosenthal2017semeval}, Semeval2018 Emotion Recognition \cite{mohammad2018semeval}, Semeval2018 Emoji Prediction challenge \cite{barbieri2018semeval}, and performed few-shot experiments with a criterion of 150 samples/category.

The experimental results in Table \ref{T5} demonstrate that our UniSA, under low-resource settings, performs well on each downstream dataset. The results reveal that UniSA has learned emotional knowledge common across subtasks during the pre-training stages and thus shows good generalization on various sentiment analysis subtasks.

%To demonstrate our proposed UniSA's generalizability on various subtasks, we perform experiments on some downstream tasks under the few-shot setting. We introduced some downstream datasets, including the SemEval2019 Hateval challenge \cite{basile-etal-2019-semeval}, Sarcasmania \cite{Siddiqui2019sarcasmania}, Semeval2017 Sentiment Analysis Challenge \cite{rosenthal2017semeval}, Semeval2018 Emotion Recognition \cite{mohammad2018semeval}, Semeval2018 Emoji Prediction challenge \cite{barbieri2018semeval}, etc., and conducted few-shot experiments with a criterion of 150 samples/category.

%As shown in Table \ref{T5}, the experimental results reveal that our UniSA (under low resource settings) performs well on each downstream dataset. It reveals that UniSA learns the emotional knowledge common across subtasks in the pre-training stages and thus has good generalization on various sentiment analysis subtasks.

\subsection{Limitation Discussion}
Error analysis in Appendix \footnote{Please see Appendix A.2 for details.} reveals that the subjective bias among datasets is one of the factors limiting UniSA's performance. However, our experiments also suggest that there are other reasons for UniSA's limitations. One of these reasons is the performance limitations of the backbone models, which can significantly improve with more parameters and learned data \cite{wei2022emergent}. Additionally, the lack of multimodal datasets poses another challenge for UniSA's performance. To address this, we encourage researchers to add more baseline datasets into SAEval and promote the development of multi-task unified modeling for sentiment analysis.
%The bias analysis in the previous section reveals that the subjective bias among datasets is one of the reasons for the limitations of UniSA. 

%The emergent abilities of large language models reveals that the performance of the models on various complex tasks can be significantly improved when the number of parameters and the learned data exceeds a certain threshold \cite{wei2022emergent}. Based on extensive experimental results, the limited performance of the backbone is another reason for the limited model performance. 

%Besides, the lack of multimodal datasets also contributes to the performance limitations of UniSA. We encourage researchers to add more baseline datasets into SAEval, to promote the development of multi-task unified modeling for sentiment analysis.

\section{Conclusion and Future Work}
In this paper, we have proposed a new benchmark for Sentiment Analysis Evaluation, SAEval, and developed a multimodal generative framework named UniSA to recouple various subtasks of sentiment analysis. To overcome the challenges of unifying multi-tasks, we have introduced the Task Specifical Prompt method and proposed novel pre-training tasks and training methods to improve the model's multimodal sentiment perception ability. Our extensive experiments have demonstrated the good generalizability of UniSA. We have also analyzed the bias between datasets and identified the limited performance of unified modeling for sentiment analysis subtasks.

In the future, we plan to expand our experiments by applying UniSA to more benchmark datasets and emotion-related tasks. We also intend to explore larger architectures as backbones for UniSA and introduce more pre-training data to improve its performance in various sentiment analysis tasks. Our work represents a step forward in the unified modeling of sentiment analysis subtasks, and we hope that it will inspire future research in this direction.

 %In this paper, we recouple the various sentiment analysis subtasks and organize the benchmark datasets of the main subtasks into a new Sentiment Analysis Evaluation benchmark, SAEval. To address the challenges faced in unifying multitasks for sentiment analysis, we propose Task Specifical Prompt method to jointly model subtasks and introduce a multimodal generative framework named UniSA to unify all multimodal subtasks as generative tasks. We propose novel pre-training tasks and training methods to improve the model's multimodal sentiment perception ability. We conduct extensive experiments and the results show that our model has good generalizability. In addition, we explore the bias between datasets and further analyze the reasons for the limited performance of unified modeling for sentiment analysis subtasks. Our work is a step forward in the unified modeling of sentiment analysis subtasks.

%In the future, we plan to apply UniSA to more benchmark datasets and emotion-related tasks. In addition, we will introduce larger architectures as the backbones of UniSA and introduce more pre-training data to improve the performance of UniSA in various sentiment analysis tasks.

\section*{Acknowledgments}
This work was supported by the following Grants: Alibaba Research Intern Program, and the National Science Foundation of China (No. 62006142). We would like to thank Tianshu Yu for his constructive comments.

\bibliographystyle{ACM-Reference-Format}
\balance
\bibliography{sample-base}

\clearpage

\appendix
\section{Appendix}
\subsection{Samples of SAEval Benchmark}
In the SAEval Benchmark, all data is formatted into the dictionary, here are some examples:
\begin{figure}[h]
  \centering
  \includegraphics[width=1\linewidth]{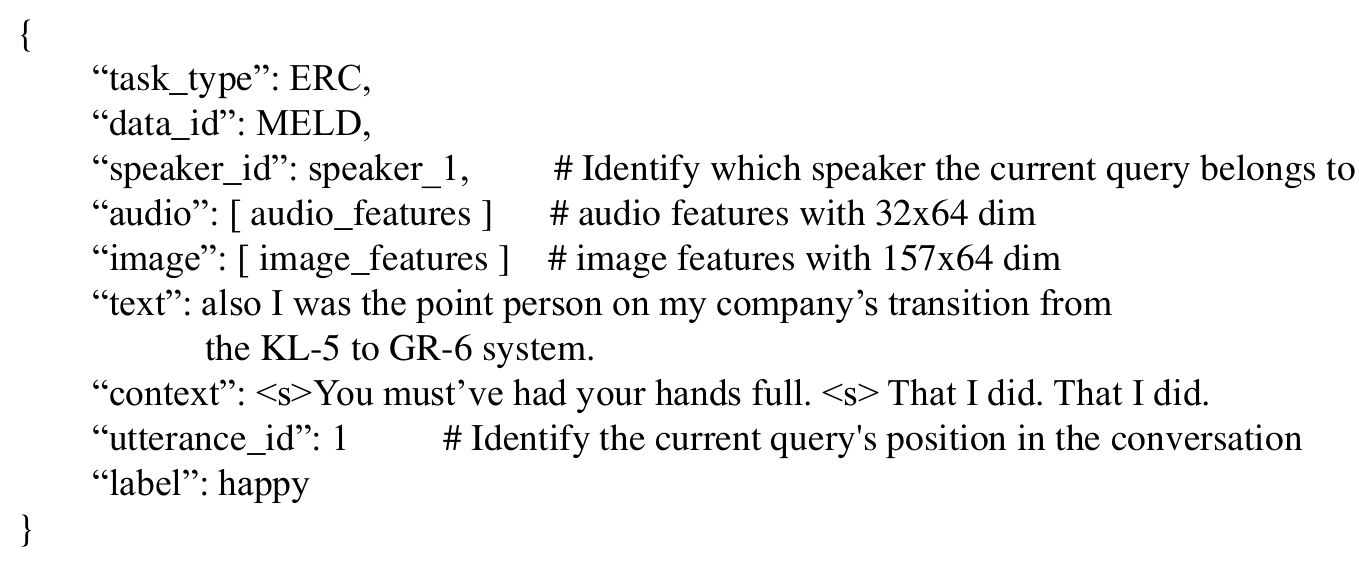}
  \vspace{-2ex}
  \caption{An example of the MELD data in SAEval.}
  \label{F5}
\end{figure}
\begin{figure}[h]
  \centering
  \includegraphics[width=1\linewidth]{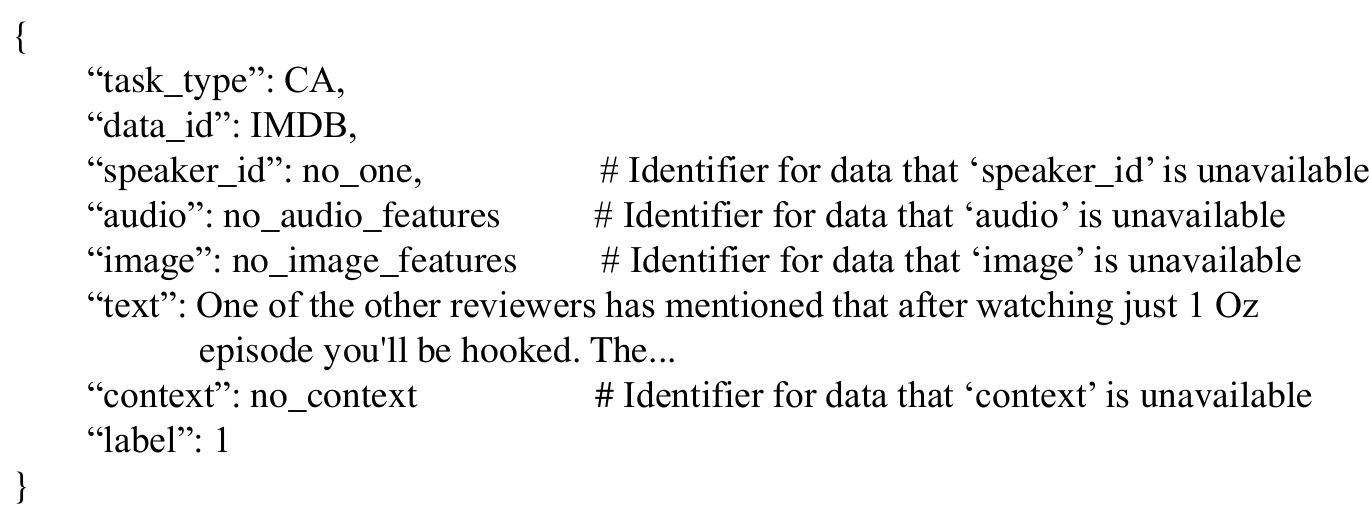}
  \caption{An example of the IMDB data in SAEval.}
  \label{F6}
\end{figure}

Clik this link for more details: https://github.com/dawn0815/SAEval-Benchmark
\subsection{Error Analysis}
In this section, we further conducted experiments on several datasets to analyze the subjective bias of the datasets from both intra-task and inter-task perspectives.
%The above experimental results reveal the excellent generalization of UniSA. However, it does not outperform SOTA on all tasks as expected, i.e., it is a "Jack of all trades, master of none."

%In this section, we further conducted experiments on some datasets to analyze the subjective bias of the datasets from both intra-task and inter-task perspectives. 
We defined the annotation bias as the difference between the accuracy of a dataset under two different annotation systems, denoted as $Bias_{ana}$:
%Given a dataset with accuracy $ACC_A$ under annotation system A and  accuracy $ACC_B$ under annotation system B, we define the annotation bias as:
\begin{equation}
Bias_{ana}=\left | ACC_{A} - ACC_{B} \right |, 
\end{equation}
where $ACC_A$ denotes the accuracy under annotation system A and $ACC_B$ refers to under accuracy under annotation system B. When there is no subjective bias between datasets $D_i$ and $D_j$, the annotation bias of these two datasets should be approximately equal under both annotation systems, i.e., $Bias_{ana}^{(i)} \simeq Bias_{ana}^{(j)}$.

The subjective bias between datasets $D_i$ and $D_j$ can be formulated as:
\begin{equation}
\label{E9}
Bias_{sub}=\left | \left | ACC_{A}^{(i)} - ACC_{B}^{(i)} \right | - \left | ACC_{A}^{(j)} - ACC_{B}^{(j)} \right | \right |,
\end{equation}
where $ACC_{A}^{(i)}$ denotes the accuracy of dataset $D_i$ under annotation system A, and $ACC_{B}^{(j)}$ denotes the accuracy of dataset $D_j$ under annotation system B.

%When there is no subjective bias between the two datasets $D_i$ and $D_j$, the annotation bias of these two datasets should be approximately equal under the annotation system A and the annotation system B:
%\begin{equation}
%Bias_{ana}^{(i)} \simeq Bias_{ana}^{(j)}
%\end{equation}
To explore the existence of subjective bias, our experiment includes the following steps: 1) Obtain the representation of each sample in datasets $D_i \in $ $\{ IEMOCAP, MELD, MOSI, EmoryNLP \}$ from UniSA's encoder output. 2) Cluster samples in dataset $D_i$ based on their real label. 3) For each sample $s$ in $D_i$, calculate its distance to each cluster (category) in other datasets $D_j \in  \{IEMOCAP, MELD, $ $MOSI,EmoryNLP\} $, where $D_i \ne D_j$. 4) Select the cluster with the smallest distance as the pseudo-label of sample $s$ under the annotation system $D_j$. 5) Generate emotion categories of dataset $D_i$ under the labeling system of dataset $D_j$ and calculate the accuracy through pseudo-labels.

The experimental results are shown in Table \ref{T6}. According to Eqn. (\ref{E9}), we can calculate the subjective bias of IEMOCAP with MELD, EmoryNLP, and MOSI, which are 20.01\%, 43.58\%, and 23.57\%, respectively. The subjective bias of MELD with EmoryNLP and MOSI is 19.1\% and 10.47\%, respectively, while the subjective bias of EmoryNLP with MOSI is 8.93\%. These experimental results reveal intra-task subjective bias (IEMOCAP, MELD, and EmoryNLP) and inter-task subjective bias (MOSI with the three other datasets).

\subsection{Training Details}
All experiments are conducted with 8 x NVIDIA RTX V100 32G. The pre-training stage one took about 3 days, due to training millions of Amazon Reviews. The pre-training stage two and finetune took about 1 day, while the few-shot learning took only minutes to complete.
\begin{table}[t]
\caption{Experimental results of subjective bias between datasets. $ACC_{iem}$, $ACC_{meld}$, $ACC_{emo}$, $ACC_{mosi}$ denote the accuracy under IEMOCAP, MELD, EmoryNLP, MOSI annotation systems, respectively.}
\label{T6}
\begin{tabular}{c|cccc}
\hline
  & $ACC_{iem}$ & $ACC_{meld}$  & $ACC_{emo}$ & $ACC_{mosi}$  \\ \hline
IEMOCAP  & 64.30   & 37.36 & 20.34 & 23.67 \\
MELD     & 55.36   & 62.29 & 37.31 & 48.12 \\
EmoryNLP & 33.88   & 28.38 & 34.26 & 26.28 \\
MOSI     & 31.48   & 23.90 & 31.63 & 48.54     \\ \hline
\end{tabular}
\end{table}

\end{document}